\documentclass[sigconf]{acmart}

\usepackage{booktabs} 
\usepackage{subfigure}
\usepackage{amsmath}
\usepackage{multirow}

\setcopyright{rightsretained}

\acmDOI{10.475/123_4}

\acmISBN{123-4567-24-567/08/06}

\acmConference[WOODSTOCK'97]{ACM Woodstock conference}{July 1997}{El
  Paso, Texas USA}
\acmYear{1997}
\copyrightyear{2016}

\acmArticle{4}
\acmPrice{15.00}

\begin{document}
\title{Linked Recurrent Neural Networks}
\titlenote{Produces the permission block, and
  copyright information}
\subtitle{Extended Abstract}
\subtitlenote{The full version of the author's guide is available as
  \texttt{acmart.pdf} document}

\author{Zhiwei Wang}
\affiliation{%
  \institution{Michigan State University}
}
\email{wangzh65@msu.edu}

\author{Yao Ma}
\affiliation{%
  \institution{Michigan State University}
}
\email{mayao4@msu.edu}

\author{Dawei Yin}
\affiliation{%
  \institution{JD.com}
}
\email{yindawei@acm.org}

\author{Jiliang Tang}
\affiliation{%
  \institution{Michigan State University}
}
\email{tangjili@msu.edu}

\begin{abstract}
Recurrent Neural Networks (RNNs) have been proven to be effective in modeling sequential data and they have been applied to boost a variety of tasks such as document classification, speech recognition and machine translation. Most of existing RNN models have been designed for sequences assumed to be identically and independently distributed (i.i.d). However, in many real-world applications, sequences are naturally linked. For example, web documents are connected by hyperlinks; and genes interact with each other. On the one hand, linked sequences are inherently not i.i.d., which poses tremendous challenges to existing RNN models. On the other hand, linked sequences offer link information in addition to the sequential information, which enables unprecedented opportunities to build advanced RNN models. In this paper, we study the problem of RNN for linked sequences. In particular, we introduce a principled approach to capture link information and propose a linked Recurrent Neural Network (LinkedRNN), which models sequential and link information coherently. We conduct experiments on real-world datasets from multiple domains and the experimental results validate the effectiveness of the proposed framework. 
\end{abstract}

%
%



\maketitle

\section{Introduction}
Recurrent Neural Networks (RNNs) have been proven to be powerful in learning a reusable parameters that produce hidden representations of sequences. They have been successfully applied to model sequential data and achieve state-of-the-art performance in numerous domains such as speech recognition~\cite{graves-etal2013,miao-etal2015,soltau-etal2016}, natural language processing~\cite{kim-etal2016,bahdanau-etal2014,mikolov-etal2010,mikolov-etal2011}, healthcare~\cite{jagannatha-etal2016,che-etal2018,luo2017}, recommendations~\cite{zhou-etal2018,hidasi-etal2015,wu-etal2016} and information retrieval~\cite{palangi-etal2016}. 

\begin{figure}
	\centering
	\includegraphics[scale=0.4]{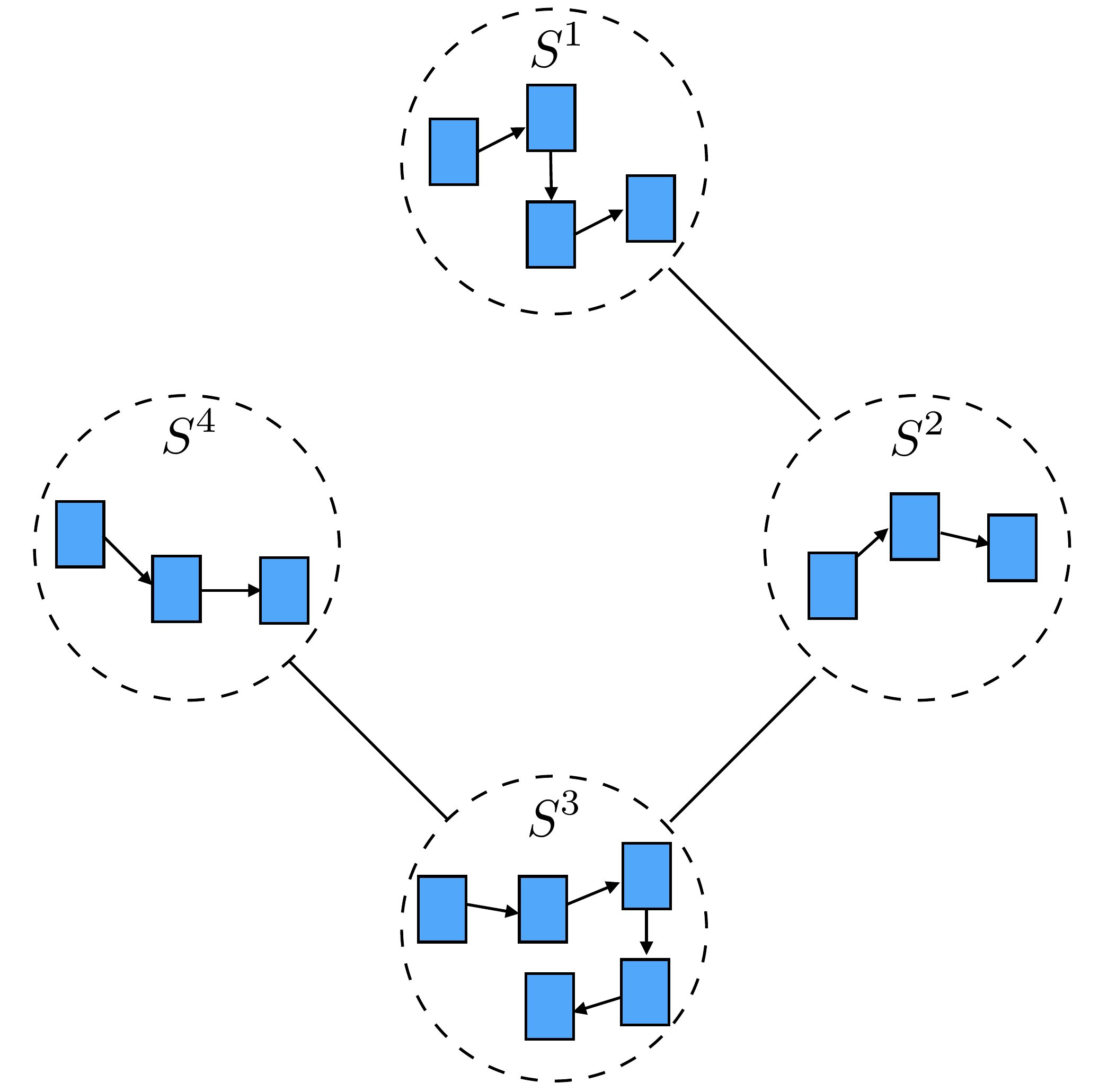}
	\caption{An Illustration of Linked Sequences. $S^1$, $S^2$, $S^3$ and $S^4$ denote four sequences and they are connected via four links.}
	\label{fig:example}
\end{figure}

The majority of existing RNN models have been designed for traditional sequences, which are assumed to be identically, independently distributed (i.i.d.). However, many real-world applications generate linked sequences. For example, web documents, sequences of words, are connected via hyperlinks; genes, sequences of DNA or RNA, typically interact with each other. Figure~\ref{fig:example} illustrates one toy example of linked sequences where there are four sequences -- $S^1$, $S^2$, $S^3$ and $S^4$. These four sequences are linked via three links -- $S^2$ is connected with $S^1$ and $S^3$ and $S^3$ is linked with $S^2$ and $S^4$. On the one hand, these linked sequences are inherently related. For example, linked web documents are likely to be similar~\cite{glover-etal2002} and interacted genes tend to share similar functionalities~\cite{bebek2012}.  Hence, linked sequences are not i.i.d., which presents immense challenges to traditional RNNs.  On the other hand, linked sequences offer additional link information in addition to the sequential information. It is evident that link information can be exploited to boost various analytical tasks such as social recommendations~\cite{tang-etal2013} , sentiment analysis~\cite{wang-etal2011,hu-etal2013} and feature selection~\cite{tang-etal2012}. Thus, the availability of link information in linked sequences has the great potential to enable us to develop advanced Recurrent Neural Networks. 

Now we have established that -- (1) traditional RNNs are insufficient and dedicated efforts are needed for linked sequences; and (2) the availability of link information in linked sequences offer unprecedented opportunities to advance traditional RNNs. In this paper, we study the problem of modeling linked sequences via RNNs. In particular, we aim to address the following challenges -- (1) how to capture link information mathematically and (2) how to combine sequential and link information via Recurrent Neural Networks. To address these two challenges, we propose a novel Linked Recurrent Neural Network (LinkedRNN) for linked sequences. Our major contributions are summarized as follows:
\begin{itemize}
	\item We introduce a principled way to capture link information for linked sequence mathematically;
	\item We propose a novel RNN framework LinkedRNN, which can model sequential and link information coherently for linked sequences; and
	\item We validate the effectiveness of the proposed framework on real-world datasets across different domains.
\end{itemize}
The rest of the paper is organized as follows. Section 2 gives a formal definition of the problem we aim to investigate. In Section 3, we motivate and detail the framework LinkedRNN. The experiment design, results and the datasets are described in Section 4. Section 5 briefly reviews the related work in literature. Finally, we conclude our work and discuss the future work in Section 6.

\section{Problem Statement}
Before we give a formal definition of the problem, we want firstly give notations that will be used throughout the paper. We denote scalars by lower-case letters such as $i$ and $j$,  vectors are denoted by bold lower-case letters such as ${\bf x}$ and ${\bf h}$, and matrices are represented by bold upper case letters such as ${\bf W}$ and ${\bf U}$. For a matrix ${\bf A}$, we denote the entry at the $i^{th}$ row and $j^{th}$ column of it as ${\bf A}(i,j)$, the $i^{th}$ row as ${\bf A}(i,:)$ and $j^{th}$ column as ${\bf A}(:,j)$. In addition, let $\{\cdots\}$ represent set where the order of the elements does not matter and the superscripts are used to denote indexes of elements such as $\{S^1, S^2, S^3\}$, which is equivalent to $\{S^3, S^2, S^1\}$. In contrast, $(\cdots)$ is used to denote a set of sequential events  where the order matters and we use subscripts to indicate the order indexes of the events in sequences such as $(x_1, x_2, x_3)$. 

Let $\mathcal{S} = \{S^1, S^2, \cdots, S^N\}$ be the set of $N$ sequences. For linked sequences, two types of information are available. One is the sequential information for each sequence. We denote the sequential information of $S^i$ as $ = (x^i_1, x^i_2, \cdots, x^i_{N^i})$ where $N^i$ is the length of $S^i$. The other is the link information. We use an adjacent matrix ${\bf A} \in \mathbb{R}^{N \times N}$ to denote the link information of linked sequences where ${\bf A}(i,j) = 1$ if there is a link between the sequence $S^i$ and $S^j$ and ${\bf A}(i,j) = 0$, otherwise. In this work, we following the transductive learning setting. In detail, we assume that a part of the sequences from $S^1$ to $S^K$ are labeled where $K < N$. We denote the labeled sequences as $\mathcal{S}_L = \{S^1, S^2, \ldots, S^K\}$. For a sequence $S^j \in \mathcal{S}_L$, we use $y^j$ to denote its label where $y^j$ is a continuous number for the regression problem and $y^j$ is one symbol for the classification problem. Note that in this work, we focus on the unweighted and undirected links among sequences. However, it is straightforward to extend the proposed framework for weighted and directed links. We would like to leave it as one future work. Although the proposed framework is designed for transductive learning, we also can use it for inductive learning, which will be discussed when we introduce the proposed framework in the following section.  

With the above notations and definitions, we formally define the problem we target in this work as follows:

{\it Given a set of sequences $\mathcal{S}$ with sequential information $S^i = (x^i_1, x^i_2, \cdots, x^i_{N^i})$ and link information ${\bf A}$, and a subset of labeled sequences $\{\mathcal{S}_L, {(y^j)}_{j=1}^K\}$, we aim to build a RNN model by leveraging $\mathcal{S}$, ${\bf A}$ and $\{\mathcal{S}_L, {(y^j)}_{j=1}^K\}$, which can learn representations for sequences to predict the labels of the unlabeled sequences in $\mathcal{S}$ }.




\section{The proposed framework}

\begin{figure*}[h]
	\centering
	\includegraphics[scale=0.62]{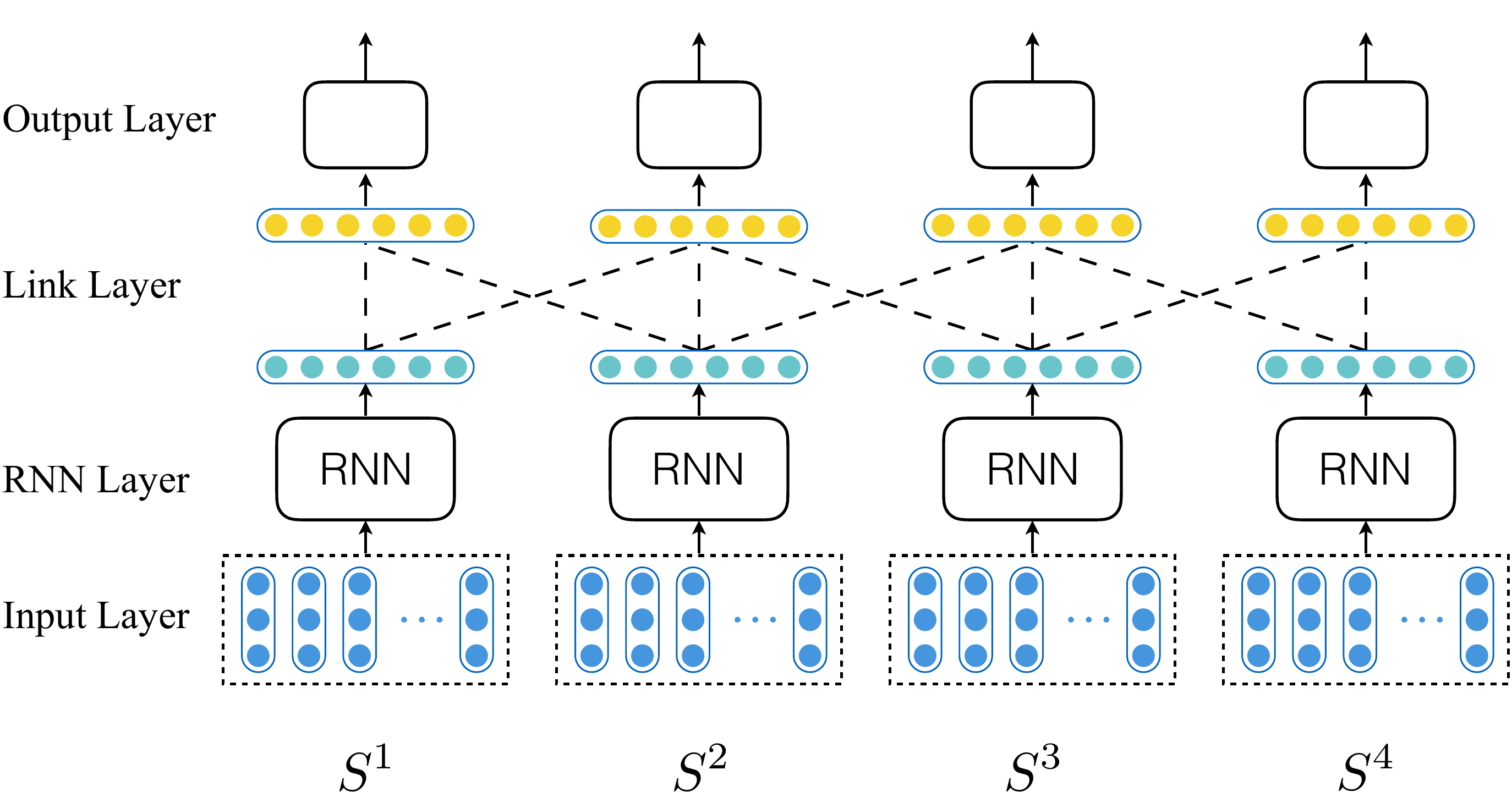}
	\caption{An illustrate of the proposed framework LinkedRNN on the toy example as shown in Figure~\ref{fig:example}. It consists of two major layers where RNN layer is to capture sequential information and the link layer is to capture link information. }
	\label{fig:model}
\end{figure*}

In addition to sequential information, link information is available for linked sequences as shown in Figure~\ref{fig:example}. As aforementioned, the major challenges to model linked sequences are how to capture link information and how to combine sequential and link information coherently. To tackle these two challenges, we propose a novel Recurrent Neural Networks LinkedRNN. An illustrate of the proposed framework on the toy example of Figure~\ref{fig:example} is demonstrated in Figure~\ref{fig:model}. It mainly consists of two layers. The RNN layer is to capture the sequential information. The output of the RNN layer is the input of the link layer where link information is captured. Next, we first detail each layer and then present the overall framework of LinkedRNN.

\subsection{Capturing sequential information}
Given a sequence $S^i = {x_1^i, x_2^i, \cdots x_{N^i}^i}$, the RNN layer aims to learn a representation vector that can capture its complex sequential patterns via Recurrent Neural Networks. In deep learning community, Recurrent Neural Networks (RNNs)\cite{rumelhart1986learning,mikolov2010recurrent} have been very successful to  capture sequential patterns in many fields\cite{mikolov2010recurrent,sutskever2011generating}. Specifically, RNN consists of recurrent units that take the previous state ${\bf h}_{t-1}^i$ and current event ${\bf x}_t^i$ as input and output a current state ${\bf h}_{t}^i$ containing the sequential information seen so far as:
\begin{align}
\label{eq:vanilla_rnn}
{\bf h}_t^i = f ({\bf U}{\bf h}_{t-1}^i + {\bf W}{\bf x}_t^i)
\end{align}
Where ${\bf U}$ and {\bf W} are the learnable parameters and $f (\cdot)$ is a activation function which enables the non-linearity. However, one major limitation of the vanilla RNN in Equation~\ref{eq:vanilla_rnn} is that it suffers from gradients vanishing or exploding issues, which fail the learning procedure as it cannot capture the error signals during back-propagation process\cite{bengio1994learning}.

More advanced recurrent units such as long short-term memory (LSTM) model~\cite{hochreiter1997long} and the Gated Recurrent Unit (GRU)~\cite{cho2014learning} have been proposed to solve the gradient vanishing problem. Different from vallina RNN, these variants employ gating mechanism to decide when and how much the state should be updated with the current information. In this work, due to its simplicity and effectiveness, we choose GRU as our RNN unit. Specifically, in the GRU, current state ${\bf h}_t^i$ is a linear interpolation between previous state ${\bf h}_{t-1}^i$ and a candidate state ${\bf \tilde{h}}_t^i$:
\begin{align}
{\bf h}_t^i =  z_t^i \odot {\bf h}_{t-1}^i + (1-z_t^i) \odot {\bf \tilde{h}}_t^i
\end{align}
\noindent where $\odot$ is the element-wise multiplication and $z_t$ is called update gate which is introduced to control how much current state should be updated.  It is obtained through the following equation:
\begin{align}
	z_t^i = \sigma({\bf W}_{z}{\bf x}_t^i + {\bf U}_{z}{\bf h}_{t-1}^i ) 
\end{align}
\noindent Where ${\bf W}_{z}$ and ${\bf U}_{z}$ are the parameters and $\sigma(\cdot)$ is the sigmoid function, that is, $\sigma(x) = \frac{1}{1 + e^{-x}}$. In addition, the newly introduced candidate state $ {\bf \tilde{h}_t}^i$ is computed by the Equation~\ref{eq:candidate_h}:
\begin{align}
	\label{eq:candidate_h}
	\tilde{h}_t^i=  g({\bf W}{\bf x}_t^i + {\bf U} (r_t^i \odot {\bf h}_{t-1}^i)) 
\end{align}
\noindent where $g(\cdot)$ is the tanh function that $g(x) = \frac{e^x - e^{-x}}{e^x + e^{-x}}$  and ${\bf W}$ and ${\bf U}$ are model parameters. $r_t$ is the reset gate which determines the contribution of previous state to the candidate state and is obtained as follows:
\begin{align}
	r_t^i =  \sigma({\bf W}_{r}{\bf x}_t^i + {\bf U}_{r}{\bf h}_{t-1}^i ) 
\end{align}

The output of the RNN layer will be the input of the link layer. For a sequence $S^i$, the RNN layer will learn a sequence of latent representations $( {\bf h}_1^i, {\bf h}_2^i, \ldots, {\bf h}_{N^i}^i)$. There are various ways to obtain the final output ${\hat {\bf h}}_i$ of $S^i$ from $( {\bf h}_1^i, {\bf h}_2^i, \ldots, {\bf h}_{N^i}^i)$. In this work, we investigate two popular ways: 
\begin{itemize}
\item As the last latent representation ${\bf h}^i_{N^i}$ is able to capture information from previous states, we can just use it as the representation of the whole sequence. We denote this way of aggregation as $aggregation_{11}$. Specifically, we let ${\hat {\bf h}_i} = {\bf h}^i_{N^i}$.
\item The attention mechanism can help the model automatically focus on relevant parts of the sequence to better capture the long-range structure and it has shown effectiveness in many tasks~\cite{bahdanau-etal2016,luong-etal2015,chorowski-etal2015}. Thus, we define our second way of aggregation based on the attention mechanism as follows:
\begin{align}
{\hat {\bf h}_i} = \sum_{j=1}^{N^i} a_j  {\bf h}^i_{j}
\end{align}
\noindent where $a_j$ is the attention score, which can be obtained as 
\begin{align}
a_j = \frac{e^{a({\bf h}^i_j)}}{\sum_m e^{a({\bf h}_m^i)}}
\end{align}
where $a({\bf h}_j^i)$ is a feedforward layer:
\begin{align}
a({\bf h}_j^i)= {\bf v}^T_a tanh({\bf W}_a {\bf h}_j^i)
\end{align}
Note that different attention mechanisms can be used, we will leave it as one future work. We denote the aggregation way described above as $aggregation_{12}$. 
\end{itemize}

For the general purpose, we will use RNN to denote GRU in the rest of the paper. 

\subsection{Capturing link information}

The RNN layer is able to capture the sequential information. However, in linked sequences, sequences are naturally related. The Homophily theory suggests that linked entities tend to have similar attributes~\cite{mcpherson-etal2001}, which have been validated in many real-world networks such as social networks~\cite{krivitsky-etal2009}, web networks~\cite{lin-etal2006}, and biological networks~\cite{bebek2012}. As indicated by Homophily, a node is likely to share similar attributes and properties with nodes with connections. In other words, a node is similar to its neighbors. With this intuition, we propose the link layer to capture link information in linked sequences. 

As shown in Figure~\ref{fig:model}, to capture link information, for a node, the link layer not only includes information from its sequential information but also aggregates information from its neighbors. The link layer can contain multiple hidden layers. In other words, for one node, we can aggregate information from itself and its neighbors multiple times. Let ${\bf v}_i^k$ be the hidden representations of the sequence $S^i$ after $k$ aggregations. Note that when $k=0$, ${\bf v}_i^0$ is the input of the link layer, i.e.,  ${\bf v}_i^0 = {\hat {\bf h}}_i$. Then ${\bf v}_i^{k+1}$ can be updated as: 
\begin{align}
\label{eq:individual_aggre}
{\bf v}_i^{k+1} = act(\frac{1}{|\mathcal{N}(i)|+1} ( {\bf v}_i^k + \sum_{S^j \in \mathcal{N}(i)} {\bf v}_j^k))
\end{align}
\noindent where $act()$ is an element-wise activation function, $\mathcal{N}(i)$ is the set of neighbors who are linked with $S^i$, i.e., $\mathcal{N}(i) = \{S^j | {\bf A}(i,j) = 1\}$, and $|\mathcal{N}(i)|$ is the number of neighbors of $S^i$. We define ${\bf V}^k = [{\bf v}_1^k, {\bf v}_2^k, \ldots, {\bf v}_N^k]$ as the matrix form of representations of all sequences at the $k$-th layer. We modify the original adjacency matrix ${\bf A}$ by allowing ${\bf A}(i,i) = 1$. The aggregation in the Eq.~(\ref{eq:individual_aggre}) can be written in the matrix form as:
\begin{align}
{\bf V}^{k+1} = act({\bf A}{\bf D}^{-1} {\bf V}^k)
\end{align}
\noindent where ${\bf V}^{k+1}$ is the embedding matrix after $k+1$ step aggregation, and ${\bf D}$ is the diagonal matrix where ${\bf D}(i,i)$ is defined as:  
\begin{align}
{\bf D}(i,i) = \sum_{j=1}^N {\bf A}(i,j)
\end{align}

\subsection{Linked Recurrent Neural Networks}

With the model components to capture sequential and link information, the procedure of the proposed framework LinkedRNN is presented below:
\begin{align}
&( {\bf h}_1^i, {\bf h}_2^i, \ldots, {\bf h}_{N^i}^i) = RNN ( S^i) \nonumber \\
& {\hat {\bf h}}_i = aggregation1 ( {\bf h}_1^i, {\bf h}_2^i, \ldots, {\bf h}_{N^i}^i) \nonumber \\
&{\bf v}_i^0 = {\hat {\bf h}}_i \nonumber \\
&{\bf v}_i^{k+1} = act(\frac{1}{|\mathcal{N}(i)|+1} ({\bf v}_i^k + \sum_{S^j \in \mathcal{N}(i)} {\bf v}_j^k)) \nonumber \\
& {\bf z}_i = aggregation2({\bf v}_i^0, {\bf v}_i^1,\ldots, {\bf v}_i^M)
\label{eq:feature-learning}
\end{align}
\noindent where the input of the RNN layer is the sequential information and the RNN layer will produce the sequence of latent representations $( {\bf h}_1^i, {\bf h}_2^i, \ldots, {\bf h}_{N^i}^i)$. The sequence of latent representations will be aggregated to obtain the output of the RNN layer, which serves as the input of the Link layer.  After $M$ layers, link layer produces a sequence of latent representations $({\bf v}_i^0, {\bf v}_i^1,\ldots, {\bf v}_i^M)$, which will be aggregated to the final representation. 

The final representation ${z}_i$ for the sequence $S^i$ is to aggregate the sequence $({\bf v}_i^0, {\bf v}_i^1,\ldots, {\bf v}_i^M)$ from the link layer. In this work, we investigate several ways to obtain the final representation ${\bf z}_i$ as:
\begin{itemize}
\item As ${\bf v}_i^M$ is the output of the last layer,  we can define the final representation as: ${\bf z}_i = {\bf v}_i^M$, and we denote this way as $aggregation_{21}$. 
\item  Although the new representation ${\bf v}^{M}$ incorporates all the neighbor information, the signal in the representation of itself may be overwhelmed during the aggregation process. This is especially likely to happen when there are a large number of neighbors. Thus, to make the new representation to focus more on itself, 
we propose to use a feed forward neural network to perform the combination. We concatenate representations from the last two layers as the input of the feed forward network. We refer this aggregation method as $aggregation_{22}$.
\item Each representation ${\bf v}_i^j$ could contain its unique information, which cannot be carried in the later part. Thus, similarly, we use a feed forward neural network to perform the combination of $({\bf v}_i^1;{\bf v}_i^2;\cdots; {\bf v}_i^M)$. We refer this aggregation method as $aggregation_{23}$.
\end{itemize}

To learn the parameters of the proposed framework LinkedRNN, we need to define a loss function that depends on the specific task. In this work, we investigate LinkedRNN in two tasks -- classification and regression.    

{\bf Classification.}  The final output of a sequence $S^i$ is  $ {\bf z}_i$. We can consider ${\bf z}_i$ as features and build the classifier. In particular, the predicted class labels can be obtained through a softmax function as:
\begin{align}
p^i = softmax({{\bf W}_c {\bf z}_i}+b_c)
\end{align}
\noindent where ${\bf W}_c$ and $b_c$ are the coefficients and the bias parameters, respectively. $p^i$ is the predicted label of the sequence $S^i$. The corresponding loss function used in this paper is the cross-entropy loss. 

{\bf Regression.} For the regression problem, we choose linear regression in this work. In other words, the regression label of the sequence $S^i$ is predicted as:
\begin{align}
p^i = {\bf W}_r {\bf z}_i +b_r
\end{align}
\noindent 
\noindent where ${\bf W}_r$ and $b_r$ are the regression coefficients and the bias parameters, respectively. Then square loss is adopted in this work as the loss function as:
\begin{align}
\label{eq:mse}
L = \frac{1}{K} \sum_{i=1}^K (y^i - p^i)^2
\end{align}

Note that there are other ways to define loss functions for classification and regression. We would like to leave the investigation of other formats of loss functions as one future work.

{\bf Prediction.} For an unlabeled sequence $S^j$ under the classification problem, its label is predicted as the one corresponding to the entity with the highest probability in $softmax({{\bf W}_c {\bf z}_j}+b_c)$. 

For an unlabeled sequence $S^j$ under the regression problem, its label is predicted as ${\bf W}_r {\bf z}_i +b_r$. 

Although the framework is designed for transductive learning, it can be naturally used for inductive learning.  For a sequence $S^k$, which is unseen in the given linked sequences $\mathcal{S}$, according to its sequential information and its neighbors $\mathcal{N}(k)$, it is easy to obtain its representation ${\bf z}_k$ via Eq.~(\ref{eq:feature-learning}). Then based on ${\bf z}_k$, its label can be predicted as the normal prediction step described above. 
\section{Experiment}

In this section, we present experimental details to verify the effectiveness of the proposed framework. Specifically, we validate the proposed framework on datasets from two different domains. Next, we firstly describe the datasets we used in the experiments and then compare the performance of the proposed framework with representative baselines. Lastly, we analyze the key components of LinkedRNN.

\begin{table}
	\begin{center}	
		\caption{Statistics of the datasets.}
		\label{table:stat}
		\begin{tabular}{p{3cm} | p{1.5cm} p{1.5cm}}
			\hline
			Description &  DBLP & BOOHEE  \\\hline
			\# of sequences & 47,491 & 18,229\\
			Network density (\textperthousand) & 0.13 & 0.012\\ 
            Avg length of sequences & 6.6 & 23.5 \\
			Max length of sequences & 20 & 29 \\
            \hline
		\end{tabular}
	\end{center}
\end{table}

\subsection{Datasets}

In this study, we collect two types of linked sequences. One is from DBLP where data contains textual sequences of papers. The other is from a weight loss website BOOHEE where data includes weight sequences of users. Some statistics of the datasets are demonstrated in Table~\ref{table:stat}. Next we introduce more details. 

{\bf DBLP dataset.} We constructed a paper citation network from the public available DBLP data set\footnote{https://aminer.org/citation.}\cite{Tang-etal2008}.  This dataset contains information for millions of paper from a variety of research fields. Specifically, each paper contains the following relevant information: paper id, publication venue, the id references of it and abstract. Following the similar practice in~\cite{tang-etal2015}, we only select papers from conferences in 10 largest computer science domains including {\it VCG, ACL, IP, TC, WC, CCS, CVPR, PDS , NIPS, KDD, WWW, ICSE, Bioinformatics, TCS}. We construct a sequence for each paper from their abstracts and regard their citation relationships as the link information between sequences. Specifically,  we first split the abstract into sentences and tokenize each sentence using python NLTK package. Then, we use Word2Vec~\cite{mikolov-etal2013} to embed each word into Euclidean space and for each sentence, we treat the mean of its word vectors as the sentence embedding. Thus, the abstract of each paper can be represented by a sequence of sentence embeddings. We will conduct the classification task on this dataset, i.e., paper classification. Thus, the label of each sequence is the corresponding publication venue. 

{\bf BOOHEE dataset.} This dataset is collected from one of the most popular weight management mobile applications, BOOHEE~\footnote{https:www.boohee.com}. It contains million of users who self-track their weights and interact with each other in the internal social network provided by the application. Specifically, they can follow friends, make comment to friends' post and mention (@) friends in comments or posts. The recored weights by users form sequences which contain the weight dynamic information and the social networking behaviors result in three networks that correspond to following, commenting, and mentioning interactions, respectively. Previous work~\cite{wang-etal2017} has shown a social correlation on the users' weight loss. Thus, we use these social networks as the link information for the weight sequence data. We preprocess the dataset to filter out the sequences from suspicious spam users. Moreover, we change the time granularity of weight sequence from days to weeks to remove the daily fluctuation noise. Specifically, we compute the mean value of all the recorded weights in one week and use it as the weight for that week. For networks, we combine three networks into one by adding them together and filter out weak ties. In this dataset, we will conduct a regression task of weight prediction. 
We choose the most recent weight in a weight sequence as the weight we aim to predict (or the groundtruth of the regression problem). Note that for a user, we remove all social interactions that form after the most recent weight where we want to avoid the issue of using future link information for weight prediction.

\begin{table*}
	\begin{center}	
		\caption{Performance Comparison in the DBLP dataset}
		\label{table:document_result}
        \begin{tabular}{p{2cm} | p{3cm} |p{2cm}|p{2cm} |p{2cm} |p{2cm}}
			\hline 
			\multirow{2}{*}{Measurement}& \multirow{2}{*}{Method}& \multicolumn{4}{c}{Training ratio}  \\
            \cline{3-6} 
			& &  10 \%  & 30 \%  & 50\% & 70\%  \\[0.7ex]
			\hline
        \multirow{6}{*}{Micro-F1}    &node2vec & 0.6641  & 0.6550  &  0.6688   &  0.6691  \\[0.7ex]	
          &GCN & 0.7005 & 0.7093 & 0.7110 & 0.7180 \\ 
		&RNN &   0.7686  &  0.7980   & 0.7978  &0.8025    \\ [0.7ex]
		&RNN-node2vec &   0.7940  &  0.8031   & 0.7933  &0.8114    \\ [0.5ex]	
          
            &RNN-GCN & 0.7912 & 0.8230 & 0.8255 & 0.8284 \\ 
            &LinkedRNN &  0.8146  & 0.8399  & 0.8463  & 0.8531   \\[0.7ex]
            
			\hline   
            \multirow{7}{*}{Macro-F1} &node2vec & 0.6514 & 0.6523 & 0.6513 &  0.6565 \\[0.7ex]	
             &GCN & 0.6874 & 0.6992 & 0.7004 & 0.7095 \\

			&RNN &   0.7452 & 0.7751  & 0.7754 & 0.7824  \\ [0.7ex]			
			&RNN+node2vec &   0.7734 &  0.7797  & 0.7702 & 0.7912  \\ [0.5ex]
            &RNN+GCN & 0.7642 & 0.8014 & 0.8069 & 0.8104 \\
		 &LinkedRNN &  0.7970 & 0.8249 & 0.8331 & 0.8365  \\[0.7ex]
        \hline  
		\end{tabular}
	\end{center}
\end{table*}

\begin{table*}
	\begin{center}	
		\caption{Performance Comparision in the BOOHEE dataset}
		\label{table:comparison}
		\begin{tabular}{ p{3cm} |p{2cm}|p{2cm} |p{2cm} |p{2cm}}
			\hline
			\multirow{2}{*}{Method }& \multicolumn{3}{c}{Training ratio} \\ 
            \cline{2-5}
			&  10 \%  & 30 \%  & 50\% & 70\%  \\   [0.5ex]
            \hline
            node2vec & 8.8702  & 8.8517  & 7.4744  & 7.0390\\[0.5ex]
            GCN &   8.9347 & 8.6830 & 6.7949 & 6.7278 \\[0.5ex]
			RNN &   8.6600  &  8.6048  & 7.0466 & 6.8033 \\ [0.5ex]
			RNN-node2vec &   8.4653  &  8.5944  & 7.0173 & 6.7796 \\ [0.5ex]
            RNN-GCN&   8.6286 & 8.5662 & 6.9967 & 6.7945 \\[0.5ex]
			LinkedRnn &  7.1822  &  6.3882  & 6.8416  &  6.3517\\[0.5ex]	
			\hline   
		\end{tabular}
	\end{center}
\end{table*}

\subsection{Representative baselines}
To validate the effectiveness of the proposed framework, we construct three groups of representative baselines. The first group includes the state-of-the-art network embedding methods, i.e., node2vec~\cite{grover-etal2016} and GCN~\cite{kipf2016semi},  which only capture the link information. The second group is the GRU RNN model~\cite{graves-etal2013}, which is the basic model we used in our model to capture sequential information. Baselines in the third group is to combine models in the first and second groups, which captures both sequential and link information. Next, we present more details about these baselines. 

\begin{itemize}
    \item{Node2vec~\cite{grover-etal2016}.} Node2vec is one state-of-the-art network embedding method. It learns the representation of sequences only capturing the link information in a random-walk perspective.
    \item{GCN~\cite{kipf2016semi}} It is the traditional graph convolutional graph algorithm. It is trained with both link and label information. Hence, it is different from node2vec, which is learnt with only link information and is totally independent on the task.   
	\item {RNN~\cite{graves-etal2013}.} RNNs have been widely used for modeling sequential data and achieved great success in a variety of domains. However, they tend to ignore the correlation between sequences and only focus on sequential information. We construct this baseline to show the importance of correlation information. To make the comparison fair, we employ the same recurrent unit (GRU) in both the proposed framework and this baseline.
    \item{RNN-node2vec.} The Node2vec method is able to learn representation from the link information and the RNN can do so from the sequential information. Thus, to obtain the representation of sequences that contains both link and sequential information, we concatenate the two sets of embeddings obtained from Node2vec and RNN via a feed forward neural network. 
     \item{RNN-GCN.} RNN-GCN applies a similar strategy of combining RNN and node2vec to combine RNN and GCN. 
\end{itemize}

There are several notes about the baselines.  First, node2vec does not use label information and it is unsupervised , RNN and RNN-node2vec utilize label information and they are supervised, and GCN and RNN-GCN use both label information and unlabeled data and they are semi-supervised. Second, some sequences may not have link information and baselines only capture link information cannot learn representations for these sequences; hence, in this work, when representations from link information are unavailable, we will use the representations from the sequential information via RNN instead. Third, we do not choose LSTM and its variants as baselines since our current model is based on GRU and  we also can choose LSTM and its variants as the base models. 

\subsection{Experimental settings}
{\bf Data split:} For both datasets, we randomly select 30\% for test. Then we fix the test set and choose $x\%$ of the remaining $70\%$ data for training and $1-x\%$ for validation to select parameters for baselines and the proposed framework. In this work, we vary $x$ as $\{10, 30, 50, 70\}$.

{\bf Parameter selection:} In our experiments, we set the dimension of representation vectors of sequences to 100. For Node2vec, we use the validation data to select the best value for $p$ and $q$ from $\{0.25, 0.50, 1, 2, 4\}$ as suggested by the authors~\cite{grover-etal2016} and use the default values for the remaining parameters. 
In addition, the learning rate for all of the methods are selected through validation set. 

{\bf Evaluation metrics:} Since we will perform  classification in the DBLP data,  we use Micro and Macro F1 scores as the metrics for DBLP, which are widely used for classification problems~\cite{yang-etal2016,grover-etal2016}. The higher value means better performance. 
We perform the regression problem weight prediction in the BOOHEE data. Therefore the performance in BOOHEE data is evaluated by mean squared error (MSE) score. The lower value of MSE indicates higher prediction performance.

\subsection{Experimental Results}
We first present the results in DBLP data. The results are shown in Table~\ref{table:document_result}. For the proposed framework, we choose $M=2$ for the link layer and more details about discussions about the choices of its aggregation functions will be discussed in the following section. From the table, we make the following observations:
\begin{itemize}
\item As we can see in Table~\ref{table:document_result}, in most cases, the performance tends to improve as the number of training samples increases.
\item The random guess can obtain 0.1 for both micro-F1 and macro-F1. We note that the network embedding methods perform much better than the random guess, which clearly shows that the link information is indeed helpful for the prediction.
\item GCN achieves much better performance than node2vec. As we mentioned before, GCN uses label information and the learnt representations are optimal for the given task.  While node2vec learns representations independent on the given task, the representations may be not optimal. 
\item The RNN approach has higher performance than GCN. Both of them use the label information.  This observation suggests that the content and sequential information is very helpful.
\item Most of the time, RNN-node2vec and RNN-GCN outperform the individual models. This observation indicates that both sequential and link information are important and they contain complementary information. 
\item The proposed framework LinkedRNN consistently outperforms baselines. This strongly demonstrates the effectiveness of LinkedRNN. In addition, comparing to RNN-node2vec and RNN-GCN, the proposed framework is able to jointly capture the sequential and link information coherently, which leads to significant performance gain.
\end{itemize}

We present the performance on BOOHEE in Table~\ref{table:comparison}. Overall, we make similar observations as these on DBLP as -- (1) the performance improves with the increase of number of training samples; (2) the combined models outperform individual ones most of the time and (3) the proposed framework LinkedRNN obtains the best performance. 

Via the comparison, we can conclude that both sequential and link information in the linked sequences are important and they contain complementary information. Meanwhile, the consistent impressive performance of LinkedRNN on datasets from different domains demonstrate its effectiveness in capturing the sequential and link information presented in the sequences.  

\subsection{Component Analysis}

In the proposed framework LinkedRNN, we have investigate several ways to define the two aggregation functions. In this subsection, we investigate the impact of the aggregation functions on the performance of the proposed framework LinkedRNN by defining the following variants.

\begin{itemize}
\item LinkedRNN11: it is the variant which chooses $aggregation_{11}$ and $aggregation_{21}$
\item LinkedRNN12: we define the variant by using  $aggregation_{11}$ and $aggregation_{22}$
\item LinkedRNN13: this variant is made by applying $aggregation_{11}$ and $aggregation_{23}$
\item LinkedRNN21: this variant utilizes $aggregation_{12}$ and $aggregation_{21}$
\item LinkedRNN22: it is the variant which chooses $aggregation_{12}$ and $aggregation_{22}$
\item LinkedRNN23: we construct the variant by adopting $aggregation_{12}$ and $aggregation_{23}$
\end{itemize}

\begin{figure}[h]
	\centering
	\includegraphics[scale=0.51]{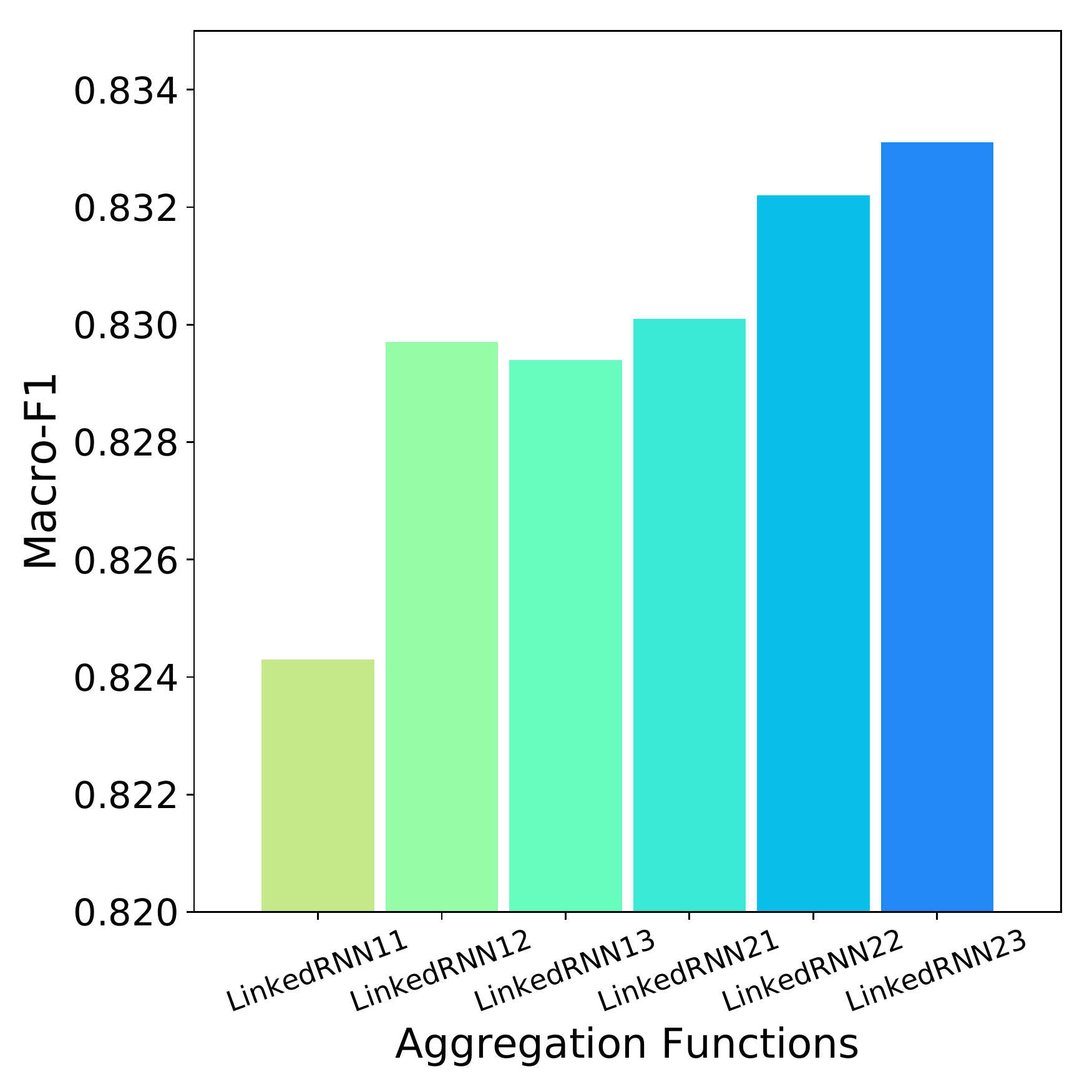}
	\caption{The impact of aggregation functions on the performance of the proposed framework.}
	\label{fig:aggregation}
\end{figure}

The results are demonstrated in Figure~\ref{fig:aggregation}. Note that we only show results on DBLP with $50\%$ as training since we can have similar observations with other settings.  It can be observed: 
\begin{itemize}
\item Generally, the variants of LinkedRNN with $aggregation_{12}$ obtain better performance than  $aggregation_{11}$. It demonstrates that aggregating the sequence of the latent presentations with the help of the attention mechanism can boost the performance. 
\item Aggregating representations from more layers in the link layer typically can result in better performance. 
\end{itemize}

\begin{figure}[h]
	\centering
	\includegraphics[scale=0.51]{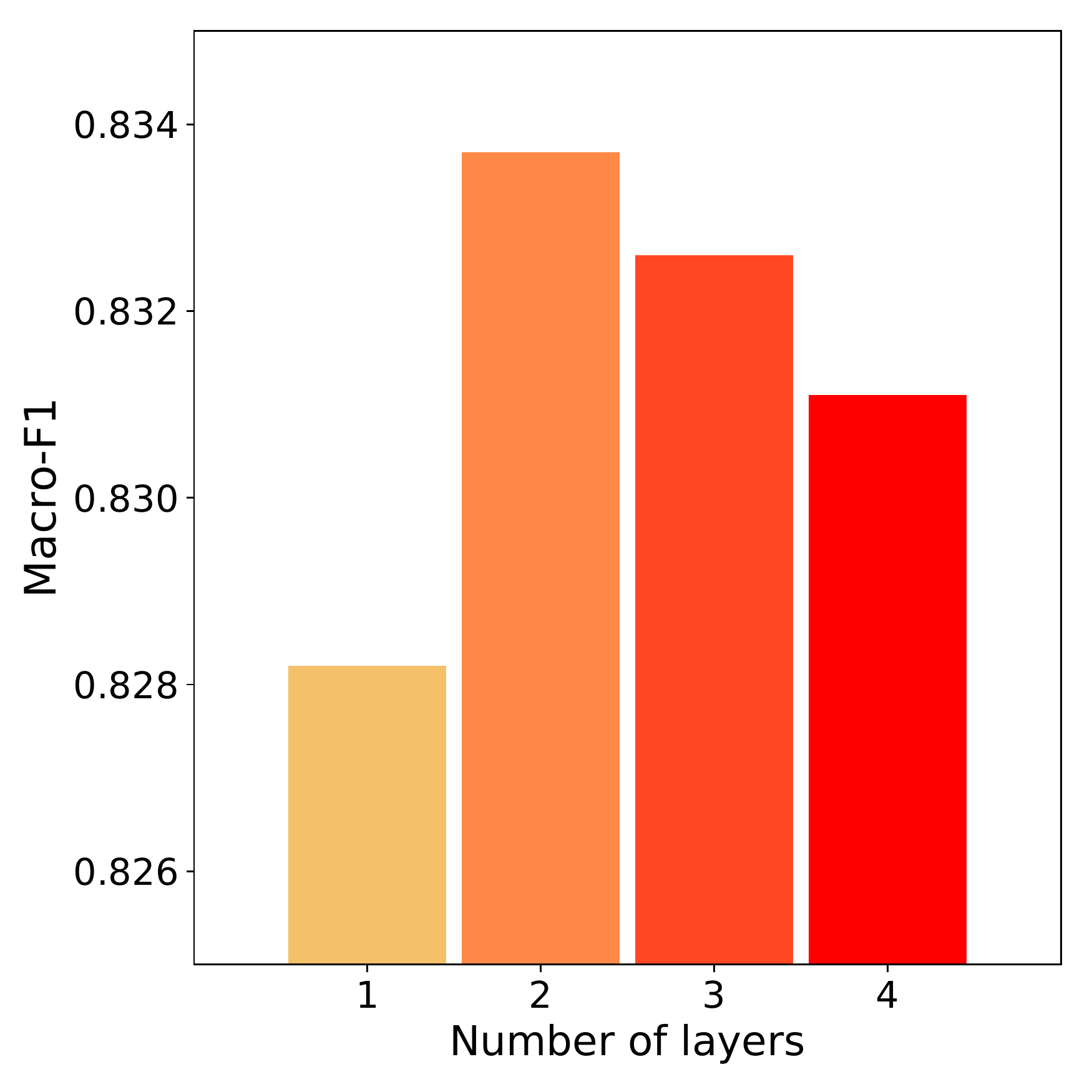}
	\caption{The performance variance with the number of layers of the link layer.}
	\label{fig:layers}
\end{figure}

\subsection{Parameter Analysis}

LinkedRNN uses the link layer to capture link information. The link layer can have multiple layers. In this subsection, we study the impact of the number of layers on the performance of LinkedRNN. The performance changes with the number of layers are shown in Figure~\ref{fig:layers}. Similar to the component analysis, we only report the results with one setting in DBLP since we have similar observations. In general, the performance first dramatically increases and then slowly decreases. One layer is not sufficient to capture the link information while more layers may result in overfitting. 
\section{Related Work}
In this section, we briefly review the RNN based deep methods that have been proposed to learn the sequential data effectively~\cite{rumelhart-etal1986}.  Although it has been designed to model arbitrarily long sequences, there are tremendous challenges to prevent it from effectively capturing the long-term dependencies. For example, the gradient vanishing and exploding issues make it very difficult to back-propagate error signals and the dependencies in sequences are complex. In addition, it is also time-consuming to train these models as the training procedure is hard to parallelize. Thus, many researchers have attempted to develop advanced architectures to overcome aforementioned challenges. One of the most successful attempts is to add gate mechanism to the recurrent unit. The two representative works are Long short-term memory (LSTM)~\cite{hochreiter-etal1997} and gated recurrent units (GRU)~\cite{cho2014learning}, where sophisticated activation function is introduced to capture long-term dependencies in sequences. For example, in LSTM, a recurrent unit maintains two states and three gates which decide how much the new memory should added, how much exiting memory should be forgotten, the amount of memory context exposure, respectively. The RNNs that are equipped with such gating mechanism can effectively mitigate the gradient exploding and vanishing issues and have demonstrated extraordinary performance in a variety of tasks, such as machine translation~\cite{sutskever-etal2014}, speech recognition~\cite{graves-etal}, and medical events detection~\cite{jagannatha-etal2016}. Moreover, several works have introduced additional gates into LSTM unit to deal with the situation where irregularly sampled data presents~\cite{neil-etal2016,baytas2017patient}. These time-aware models can largely improve the training efficiency and effectiveness of RNNs.

Beside gating mechanism, other directions of extending RNN for better performance are also heavily explored. Schuster et al~\cite{schuster-etal1997} described a new architecture called bidirectional recurrent neural networks(BRNNs), where two hidden layers that process the input from opposite directions were proposed. Although BRNNs are able to use all available input sequential information and  effectively boost the prediction performance~\cite{ma-etal2017}, the limitation of it is quite obvious as it requires the information from the future~\cite{lipton-etal2015}. Recently, Koutnik {\it et al.}~\cite{koutni-etalk2014} presented a clockwork RNN which modifies standard RNN architecture and partitions the hidden layer into separate modules. In this way, each individual can process the sequence at its own temporal granularity.  One recent work proposed by Change {\it et al.}~\cite{chang-etal2017} tries to tackle those major challenges together. In doing so, they introduced a dilated recurrent skip connection which can largely reduce the model parameters and therefore enhances the computational efficiency. In addition, such layers can be stacked so that the dependencies of different scales are learned effectively at different layers. While a large body of research has focused on modeling the dependencies within the sequences, limited efforts have been made to model the dependencies between sequences. In this paper, we devote to tackling this novel challenge brought by the links among sequences and propose an effective model which has shown promising results.   

\section{Conclusion}
RNNs have been proven to be powerful in modeling sequences in many domains. Most of existing RNN methods have been designed for sequences which are assumed to be i.i.d. However, in many real-world applications, sequences are inherently linked and linked sequences present both challenges and opportunities to existing RNN methods, which calls for novel RNN methods. In this paper, we study the problem of designing RNN models for linked sequences. Suggested by Homophily, we introduce a principled method to capture link information and propose a novel RNN framework LinkedRNN, which can jointly model sequential and link information. Experimental results on datasets from different domains demonstrate that (1) the proposed framework can outperform a variety of representative baselines; and (2) link information is helpful to boost the RNN performance.  

There are several interesting directions to investigate in the future. First, our current model focuses on unweighted and undirected links and we will study weighted and directed links and the corresponding RNN models. Second, in current work, we focus on classification and regression problems with certain loss functions. we will investigate other types of loss functions to learn the parameters of the proposed framework and also investigate more applications of the proposed framework. Third, since our model can be naturally extended for inductive learning, we will further validate the effectiveness of the proposed framework for inductive learning. Finally, in some applications, the link information may be evolving; thus we plan to study RNN models, which can capture the dynamics of links as well.

\bibliographystyle{ACM-Reference-Format}
\bibliography{zhiwei}


\begin{thebibliography}{50}


\ifx \showCODEN    \undefined \def \showCODEN     #1{\unskip}     \fi
\ifx \showDOI      \undefined \def \showDOI       #1{#1}\fi
\ifx \showISBNx    \undefined \def \showISBNx     #1{\unskip}     \fi
\ifx \showISBNxiii \undefined \def \showISBNxiii  #1{\unskip}     \fi
\ifx \showISSN     \undefined \def \showISSN      #1{\unskip}     \fi
\ifx \showLCCN     \undefined \def \showLCCN      #1{\unskip}     \fi
\ifx \shownote     \undefined \def \shownote      #1{#1}          \fi
\ifx \showarticletitle \undefined \def \showarticletitle #1{#1}   \fi
\ifx \showURL      \undefined \def \showURL       {\relax}        \fi
\providecommand\bibfield[2]{#2}
\providecommand\bibinfo[2]{#2}
\providecommand\natexlab[1]{#1}
\providecommand\showeprint[2][]{arXiv:#2}

\bibitem[\protect\citeauthoryear{Bahdanau, Cho, and Bengio}{Bahdanau
  et~al\mbox{.}}{2014}]%
        {bahdanau-etal2014}
\bibfield{author}{\bibinfo{person}{Dzmitry Bahdanau},
  \bibinfo{person}{Kyunghyun Cho}, {and} \bibinfo{person}{Yoshua Bengio}.}
  \bibinfo{year}{2014}\natexlab{}.
\newblock \showarticletitle{Neural machine translation by jointly learning to
  align and translate}.
\newblock \bibinfo{journal}{\emph{arXiv preprint arXiv:1409.0473}}
  (\bibinfo{year}{2014}).
\newblock


\bibitem[\protect\citeauthoryear{Bahdanau, Chorowski, Serdyuk, Brakel, and
  Bengio}{Bahdanau et~al\mbox{.}}{2016}]%
        {bahdanau-etal2016}
\bibfield{author}{\bibinfo{person}{Dzmitry Bahdanau}, \bibinfo{person}{Jan
  Chorowski}, \bibinfo{person}{Dmitriy Serdyuk}, \bibinfo{person}{Philemon
  Brakel}, {and} \bibinfo{person}{Yoshua Bengio}.}
  \bibinfo{year}{2016}\natexlab{}.
\newblock \showarticletitle{End-to-end attention-based large vocabulary speech
  recognition}. In \bibinfo{booktitle}{\emph{Acoustics, Speech and Signal
  Processing (ICASSP), 2016 IEEE International Conference on}}. IEEE,
  \bibinfo{pages}{4945--4949}.
\newblock


\bibitem[\protect\citeauthoryear{Baytas, Xiao, Zhang, Wang, Jain, and
  Zhou}{Baytas et~al\mbox{.}}{2017}]%
        {baytas2017patient}
\bibfield{author}{\bibinfo{person}{Inci~M Baytas}, \bibinfo{person}{Cao Xiao},
  \bibinfo{person}{Xi Zhang}, \bibinfo{person}{Fei Wang},
  \bibinfo{person}{Anil~K Jain}, {and} \bibinfo{person}{Jiayu Zhou}.}
  \bibinfo{year}{2017}\natexlab{}.
\newblock \showarticletitle{Patient subtyping via time-aware LSTM networks}. In
  \bibinfo{booktitle}{\emph{Proceedings of the 23rd ACM SIGKDD International
  Conference on Knowledge Discovery and Data Mining}}. ACM,
  \bibinfo{pages}{65--74}.
\newblock


\bibitem[\protect\citeauthoryear{Bebek}{Bebek}{2012}]%
        {bebek2012}
\bibfield{author}{\bibinfo{person}{Gurkan Bebek}.}
  \bibinfo{year}{2012}\natexlab{}.
\newblock \showarticletitle{Identifying gene interaction networks}.
\newblock In \bibinfo{booktitle}{\emph{Statistical Human Genetics}}.
  \bibinfo{publisher}{Springer}, \bibinfo{pages}{483--494}.
\newblock


\bibitem[\protect\citeauthoryear{Bengio, Simard, and Frasconi}{Bengio
  et~al\mbox{.}}{1994}]%
        {bengio1994learning}
\bibfield{author}{\bibinfo{person}{Yoshua Bengio}, \bibinfo{person}{Patrice
  Simard}, {and} \bibinfo{person}{Paolo Frasconi}.}
  \bibinfo{year}{1994}\natexlab{}.
\newblock \showarticletitle{Learning long-term dependencies with gradient
  descent is difficult}.
\newblock \bibinfo{journal}{\emph{IEEE transactions on neural networks}}
  \bibinfo{volume}{5}, \bibinfo{number}{2} (\bibinfo{year}{1994}),
  \bibinfo{pages}{157--166}.
\newblock


\bibitem[\protect\citeauthoryear{Chang, Zhang, Han, Yu, Guo, Tan, Cui,
  Witbrock, Hasegawa-Johnson, and Huang}{Chang et~al\mbox{.}}{2017}]%
        {chang-etal2017}
\bibfield{author}{\bibinfo{person}{Shiyu Chang}, \bibinfo{person}{Yang Zhang},
  \bibinfo{person}{Wei Han}, \bibinfo{person}{Mo Yu}, \bibinfo{person}{Xiaoxiao
  Guo}, \bibinfo{person}{Wei Tan}, \bibinfo{person}{Xiaodong Cui},
  \bibinfo{person}{Michael Witbrock}, \bibinfo{person}{Mark~A
  Hasegawa-Johnson}, {and} \bibinfo{person}{Thomas~S Huang}.}
  \bibinfo{year}{2017}\natexlab{}.
\newblock \showarticletitle{Dilated recurrent neural networks}. In
  \bibinfo{booktitle}{\emph{Advances in Neural Information Processing
  Systems}}. \bibinfo{pages}{76--86}.
\newblock


\bibitem[\protect\citeauthoryear{Che, Purushotham, Cho, Sontag, and Liu}{Che
  et~al\mbox{.}}{2018}]%
        {che-etal2018}
\bibfield{author}{\bibinfo{person}{Zhengping Che}, \bibinfo{person}{Sanjay
  Purushotham}, \bibinfo{person}{Kyunghyun Cho}, \bibinfo{person}{David
  Sontag}, {and} \bibinfo{person}{Yan Liu}.} \bibinfo{year}{2018}\natexlab{}.
\newblock \showarticletitle{Recurrent neural networks for multivariate time
  series with missing values}.
\newblock \bibinfo{journal}{\emph{Scientific reports}} \bibinfo{volume}{8},
  \bibinfo{number}{1} (\bibinfo{year}{2018}), \bibinfo{pages}{6085}.
\newblock


\bibitem[\protect\citeauthoryear{Cho, Van~Merri{\"e}nboer, Gulcehre, Bahdanau,
  Bougares, Schwenk, and Bengio}{Cho et~al\mbox{.}}{2014}]%
        {cho2014learning}
\bibfield{author}{\bibinfo{person}{Kyunghyun Cho}, \bibinfo{person}{Bart
  Van~Merri{\"e}nboer}, \bibinfo{person}{Caglar Gulcehre},
  \bibinfo{person}{Dzmitry Bahdanau}, \bibinfo{person}{Fethi Bougares},
  \bibinfo{person}{Holger Schwenk}, {and} \bibinfo{person}{Yoshua Bengio}.}
  \bibinfo{year}{2014}\natexlab{}.
\newblock \showarticletitle{Learning phrase representations using RNN
  encoder-decoder for statistical machine translation}.
\newblock \bibinfo{journal}{\emph{arXiv preprint arXiv:1406.1078}}
  (\bibinfo{year}{2014}).
\newblock


\bibitem[\protect\citeauthoryear{Chorowski, Bahdanau, Serdyuk, Cho, and
  Bengio}{Chorowski et~al\mbox{.}}{2015}]%
        {chorowski-etal2015}
\bibfield{author}{\bibinfo{person}{Jan~K Chorowski}, \bibinfo{person}{Dzmitry
  Bahdanau}, \bibinfo{person}{Dmitriy Serdyuk}, \bibinfo{person}{Kyunghyun
  Cho}, {and} \bibinfo{person}{Yoshua Bengio}.}
  \bibinfo{year}{2015}\natexlab{}.
\newblock \showarticletitle{Attention-based models for speech recognition}. In
  \bibinfo{booktitle}{\emph{Advances in neural information processing
  systems}}. \bibinfo{pages}{577--585}.
\newblock


\bibitem[\protect\citeauthoryear{Glover, Tsioutsiouliklis, Lawrence, Pennock,
  and Flake}{Glover et~al\mbox{.}}{2002}]%
        {glover-etal2002}
\bibfield{author}{\bibinfo{person}{Eric~J Glover}, \bibinfo{person}{Kostas
  Tsioutsiouliklis}, \bibinfo{person}{Steve Lawrence}, \bibinfo{person}{David~M
  Pennock}, {and} \bibinfo{person}{Gary~W Flake}.}
  \bibinfo{year}{2002}\natexlab{}.
\newblock \showarticletitle{Using web structure for classifying and describing
  web pages}. In \bibinfo{booktitle}{\emph{Proceedings of the 11th
  international conference on World Wide Web}}. ACM, \bibinfo{pages}{562--569}.
\newblock


\bibitem[\protect\citeauthoryear{Graves, Jaitly, and Mohamed}{Graves
  et~al\mbox{.}}{2013a}]%
        {graves-etal}
\bibfield{author}{\bibinfo{person}{Alex Graves}, \bibinfo{person}{Navdeep
  Jaitly}, {and} \bibinfo{person}{Abdel-rahman Mohamed}.}
  \bibinfo{year}{2013}\natexlab{a}.
\newblock \showarticletitle{Hybrid speech recognition with deep bidirectional
  LSTM}. In \bibinfo{booktitle}{\emph{Automatic Speech Recognition and
  Understanding (ASRU), 2013 IEEE Workshop on}}. IEEE,
  \bibinfo{pages}{273--278}.
\newblock


\bibitem[\protect\citeauthoryear{Graves, Mohamed, and Hinton}{Graves
  et~al\mbox{.}}{2013b}]%
        {graves-etal2013}
\bibfield{author}{\bibinfo{person}{Alex Graves}, \bibinfo{person}{Abdel-rahman
  Mohamed}, {and} \bibinfo{person}{Geoffrey Hinton}.}
  \bibinfo{year}{2013}\natexlab{b}.
\newblock \showarticletitle{Speech recognition with deep recurrent neural
  networks}. In \bibinfo{booktitle}{\emph{Acoustics, speech and signal
  processing (icassp), 2013 ieee international conference on}}. IEEE,
  \bibinfo{pages}{6645--6649}.
\newblock


\bibitem[\protect\citeauthoryear{Grover and Leskovec}{Grover and
  Leskovec}{2016}]%
        {grover-etal2016}
\bibfield{author}{\bibinfo{person}{Aditya Grover} {and} \bibinfo{person}{Jure
  Leskovec}.} \bibinfo{year}{2016}\natexlab{}.
\newblock \showarticletitle{node2vec: Scalable feature learning for networks}.
  In \bibinfo{booktitle}{\emph{Proceedings of the 22nd ACM SIGKDD international
  conference on Knowledge discovery and data mining}}. ACM,
  \bibinfo{pages}{855--864}.
\newblock


\bibitem[\protect\citeauthoryear{Hidasi, Karatzoglou, Baltrunas, and
  Tikk}{Hidasi et~al\mbox{.}}{2015}]%
        {hidasi-etal2015}
\bibfield{author}{\bibinfo{person}{Bal{\'a}zs Hidasi},
  \bibinfo{person}{Alexandros Karatzoglou}, \bibinfo{person}{Linas Baltrunas},
  {and} \bibinfo{person}{Domonkos Tikk}.} \bibinfo{year}{2015}\natexlab{}.
\newblock \showarticletitle{Session-based recommendations with recurrent neural
  networks}.
\newblock \bibinfo{journal}{\emph{arXiv preprint arXiv:1511.06939}}
  (\bibinfo{year}{2015}).
\newblock


\bibitem[\protect\citeauthoryear{Hochreiter and Schmidhuber}{Hochreiter and
  Schmidhuber}{1997a}]%
        {hochreiter1997long}
\bibfield{author}{\bibinfo{person}{Sepp Hochreiter} {and}
  \bibinfo{person}{J{\"u}rgen Schmidhuber}.} \bibinfo{year}{1997}\natexlab{a}.
\newblock \showarticletitle{Long short-term memory}.
\newblock \bibinfo{journal}{\emph{Neural computation}} \bibinfo{volume}{9},
  \bibinfo{number}{8} (\bibinfo{year}{1997}), \bibinfo{pages}{1735--1780}.
\newblock


\bibitem[\protect\citeauthoryear{Hochreiter and Schmidhuber}{Hochreiter and
  Schmidhuber}{1997b}]%
        {hochreiter-etal1997}
\bibfield{author}{\bibinfo{person}{Sepp Hochreiter} {and}
  \bibinfo{person}{J{\"u}rgen Schmidhuber}.} \bibinfo{year}{1997}\natexlab{b}.
\newblock \showarticletitle{Long short-term memory}.
\newblock \bibinfo{journal}{\emph{Neural computation}} \bibinfo{volume}{9},
  \bibinfo{number}{8} (\bibinfo{year}{1997}), \bibinfo{pages}{1735--1780}.
\newblock


\bibitem[\protect\citeauthoryear{Hu, Tang, Tang, and Liu}{Hu
  et~al\mbox{.}}{2013}]%
        {hu-etal2013}
\bibfield{author}{\bibinfo{person}{Xia Hu}, \bibinfo{person}{Lei Tang},
  \bibinfo{person}{Jiliang Tang}, {and} \bibinfo{person}{Huan Liu}.}
  \bibinfo{year}{2013}\natexlab{}.
\newblock \showarticletitle{Exploiting social relations for sentiment analysis
  in microblogging}. In \bibinfo{booktitle}{\emph{Proceedings of the sixth ACM
  international conference on Web search and data mining}}. ACM,
  \bibinfo{pages}{537--546}.
\newblock


\bibitem[\protect\citeauthoryear{Jagannatha and Yu}{Jagannatha and Yu}{2016}]%
        {jagannatha-etal2016}
\bibfield{author}{\bibinfo{person}{Abhyuday~N Jagannatha} {and}
  \bibinfo{person}{Hong Yu}.} \bibinfo{year}{2016}\natexlab{}.
\newblock \showarticletitle{Bidirectional RNN for medical event detection in
  electronic health records}. In \bibinfo{booktitle}{\emph{Proceedings of the
  conference. Association for Computational Linguistics. North American
  Chapter. Meeting}}, Vol.~\bibinfo{volume}{2016}. NIH Public Access,
  \bibinfo{pages}{473}.
\newblock


\bibitem[\protect\citeauthoryear{Kim, Jernite, Sontag, and Rush}{Kim
  et~al\mbox{.}}{2016}]%
        {kim-etal2016}
\bibfield{author}{\bibinfo{person}{Yoon Kim}, \bibinfo{person}{Yacine Jernite},
  \bibinfo{person}{David Sontag}, {and} \bibinfo{person}{Alexander~M Rush}.}
  \bibinfo{year}{2016}\natexlab{}.
\newblock \showarticletitle{Character-Aware Neural Language Models.}. In
  \bibinfo{booktitle}{\emph{AAAI}}. \bibinfo{pages}{2741--2749}.
\newblock


\bibitem[\protect\citeauthoryear{Kipf and Welling}{Kipf and Welling}{2016}]%
        {kipf2016semi}
\bibfield{author}{\bibinfo{person}{Thomas~N Kipf} {and} \bibinfo{person}{Max
  Welling}.} \bibinfo{year}{2016}\natexlab{}.
\newblock \showarticletitle{Semi-supervised classification with graph
  convolutional networks}.
\newblock \bibinfo{journal}{\emph{arXiv preprint arXiv:1609.02907}}
  (\bibinfo{year}{2016}).
\newblock


\bibitem[\protect\citeauthoryear{Koutnik, Greff, Gomez, and
  Schmidhuber}{Koutnik et~al\mbox{.}}{2014}]%
        {koutni-etalk2014}
\bibfield{author}{\bibinfo{person}{Jan Koutnik}, \bibinfo{person}{Klaus Greff},
  \bibinfo{person}{Faustino Gomez}, {and} \bibinfo{person}{Juergen
  Schmidhuber}.} \bibinfo{year}{2014}\natexlab{}.
\newblock \showarticletitle{A clockwork rnn}.
\newblock \bibinfo{journal}{\emph{arXiv preprint arXiv:1402.3511}}
  (\bibinfo{year}{2014}).
\newblock


\bibitem[\protect\citeauthoryear{Krivitsky, Handcock, Raftery, and
  Hoff}{Krivitsky et~al\mbox{.}}{2009}]%
        {krivitsky-etal2009}
\bibfield{author}{\bibinfo{person}{Pavel~N Krivitsky}, \bibinfo{person}{Mark~S
  Handcock}, \bibinfo{person}{Adrian~E Raftery}, {and} \bibinfo{person}{Peter~D
  Hoff}.} \bibinfo{year}{2009}\natexlab{}.
\newblock \showarticletitle{Representing degree distributions, clustering, and
  homophily in social networks with latent cluster random effects models}.
\newblock \bibinfo{journal}{\emph{Social networks}} \bibinfo{volume}{31},
  \bibinfo{number}{3} (\bibinfo{year}{2009}), \bibinfo{pages}{204--213}.
\newblock


\bibitem[\protect\citeauthoryear{Lin, Lyu, and King}{Lin et~al\mbox{.}}{2006}]%
        {lin-etal2006}
\bibfield{author}{\bibinfo{person}{Zhenjiang Lin}, \bibinfo{person}{Michael~R
  Lyu}, {and} \bibinfo{person}{Irwin King}.} \bibinfo{year}{2006}\natexlab{}.
\newblock \showarticletitle{PageSim: a novel link-based measure of web page
  aimilarity}. In \bibinfo{booktitle}{\emph{Proceedings of the 15th
  international conference on World Wide Web}}. ACM,
  \bibinfo{pages}{1019--1020}.
\newblock


\bibitem[\protect\citeauthoryear{Lipton, Berkowitz, and Elkan}{Lipton
  et~al\mbox{.}}{2015}]%
        {lipton-etal2015}
\bibfield{author}{\bibinfo{person}{Zachary~C Lipton}, \bibinfo{person}{John
  Berkowitz}, {and} \bibinfo{person}{Charles Elkan}.}
  \bibinfo{year}{2015}\natexlab{}.
\newblock \showarticletitle{A critical review of recurrent neural networks for
  sequence learning}.
\newblock \bibinfo{journal}{\emph{arXiv preprint arXiv:1506.00019}}
  (\bibinfo{year}{2015}).
\newblock


\bibitem[\protect\citeauthoryear{Luo}{Luo}{2017}]%
        {luo2017}
\bibfield{author}{\bibinfo{person}{Yuan Luo}.} \bibinfo{year}{2017}\natexlab{}.
\newblock \showarticletitle{Recurrent neural networks for classifying relations
  in clinical notes}.
\newblock \bibinfo{journal}{\emph{Journal of biomedical informatics}}
  \bibinfo{volume}{72} (\bibinfo{year}{2017}), \bibinfo{pages}{85--95}.
\newblock


\bibitem[\protect\citeauthoryear{Luong, Pham, and Manning}{Luong
  et~al\mbox{.}}{2015}]%
        {luong-etal2015}
\bibfield{author}{\bibinfo{person}{Minh-Thang Luong}, \bibinfo{person}{Hieu
  Pham}, {and} \bibinfo{person}{Christopher~D Manning}.}
  \bibinfo{year}{2015}\natexlab{}.
\newblock \showarticletitle{Effective approaches to attention-based neural
  machine translation}.
\newblock \bibinfo{journal}{\emph{arXiv preprint arXiv:1508.04025}}
  (\bibinfo{year}{2015}).
\newblock


\bibitem[\protect\citeauthoryear{Ma, Chitta, Zhou, You, Sun, and Gao}{Ma
  et~al\mbox{.}}{2017}]%
        {ma-etal2017}
\bibfield{author}{\bibinfo{person}{Fenglong Ma}, \bibinfo{person}{Radha
  Chitta}, \bibinfo{person}{Jing Zhou}, \bibinfo{person}{Quanzeng You},
  \bibinfo{person}{Tong Sun}, {and} \bibinfo{person}{Jing Gao}.}
  \bibinfo{year}{2017}\natexlab{}.
\newblock \showarticletitle{Dipole: Diagnosis prediction in healthcare via
  attention-based bidirectional recurrent neural networks}. In
  \bibinfo{booktitle}{\emph{Proceedings of the 23rd ACM SIGKDD International
  Conference on Knowledge Discovery and Data Mining}}. ACM,
  \bibinfo{pages}{1903--1911}.
\newblock


\bibitem[\protect\citeauthoryear{McPherson, Smith-Lovin, and Cook}{McPherson
  et~al\mbox{.}}{2001}]%
        {mcpherson-etal2001}
\bibfield{author}{\bibinfo{person}{Miller McPherson}, \bibinfo{person}{Lynn
  Smith-Lovin}, {and} \bibinfo{person}{James~M Cook}.}
  \bibinfo{year}{2001}\natexlab{}.
\newblock \showarticletitle{Birds of a feather: Homophily in social networks}.
\newblock \bibinfo{journal}{\emph{Annual review of sociology}}
  \bibinfo{volume}{27}, \bibinfo{number}{1} (\bibinfo{year}{2001}),
  \bibinfo{pages}{415--444}.
\newblock


\bibitem[\protect\citeauthoryear{Miao, Gowayyed, and Metze}{Miao
  et~al\mbox{.}}{2015}]%
        {miao-etal2015}
\bibfield{author}{\bibinfo{person}{Yajie Miao}, \bibinfo{person}{Mohammad
  Gowayyed}, {and} \bibinfo{person}{Florian Metze}.}
  \bibinfo{year}{2015}\natexlab{}.
\newblock \showarticletitle{EESEN: End-to-end speech recognition using deep RNN
  models and WFST-based decoding}. In \bibinfo{booktitle}{\emph{Automatic
  Speech Recognition and Understanding (ASRU), 2015 IEEE Workshop on}}. IEEE,
  \bibinfo{pages}{167--174}.
\newblock


\bibitem[\protect\citeauthoryear{Mikolov, Chen, Corrado, and Dean}{Mikolov
  et~al\mbox{.}}{2013}]%
        {mikolov-etal2013}
\bibfield{author}{\bibinfo{person}{Tomas Mikolov}, \bibinfo{person}{Kai Chen},
  \bibinfo{person}{Greg Corrado}, {and} \bibinfo{person}{Jeffrey Dean}.}
  \bibinfo{year}{2013}\natexlab{}.
\newblock \showarticletitle{Efficient estimation of word representations in
  vector space}.
\newblock \bibinfo{journal}{\emph{arXiv preprint arXiv:1301.3781}}
  (\bibinfo{year}{2013}).
\newblock


\bibitem[\protect\citeauthoryear{Mikolov, Karafi{\'a}t, Burget,
  {\v{C}}ernock{\`y}, and Khudanpur}{Mikolov et~al\mbox{.}}{2010a}]%
        {mikolov-etal2010}
\bibfield{author}{\bibinfo{person}{Tom{\'a}{\v{s}} Mikolov},
  \bibinfo{person}{Martin Karafi{\'a}t}, \bibinfo{person}{Luk{\'a}{\v{s}}
  Burget}, \bibinfo{person}{Jan {\v{C}}ernock{\`y}}, {and}
  \bibinfo{person}{Sanjeev Khudanpur}.} \bibinfo{year}{2010}\natexlab{a}.
\newblock \showarticletitle{Recurrent neural network based language model}. In
  \bibinfo{booktitle}{\emph{Eleventh Annual Conference of the International
  Speech Communication Association}}.
\newblock


\bibitem[\protect\citeauthoryear{Mikolov, Karafi{\'a}t, Burget,
  {\v{C}}ernock{\`y}, and Khudanpur}{Mikolov et~al\mbox{.}}{2010b}]%
        {mikolov2010recurrent}
\bibfield{author}{\bibinfo{person}{Tom{\'a}{\v{s}} Mikolov},
  \bibinfo{person}{Martin Karafi{\'a}t}, \bibinfo{person}{Luk{\'a}{\v{s}}
  Burget}, \bibinfo{person}{Jan {\v{C}}ernock{\`y}}, {and}
  \bibinfo{person}{Sanjeev Khudanpur}.} \bibinfo{year}{2010}\natexlab{b}.
\newblock \showarticletitle{Recurrent neural network based language model}. In
  \bibinfo{booktitle}{\emph{Eleventh Annual Conference of the International
  Speech Communication Association}}.
\newblock


\bibitem[\protect\citeauthoryear{Mikolov, Kombrink, Burget, {\v{C}}ernock{\`y},
  and Khudanpur}{Mikolov et~al\mbox{.}}{2011}]%
        {mikolov-etal2011}
\bibfield{author}{\bibinfo{person}{Tom{\'a}{\v{s}} Mikolov},
  \bibinfo{person}{Stefan Kombrink}, \bibinfo{person}{Luk{\'a}{\v{s}} Burget},
  \bibinfo{person}{Jan {\v{C}}ernock{\`y}}, {and} \bibinfo{person}{Sanjeev
  Khudanpur}.} \bibinfo{year}{2011}\natexlab{}.
\newblock \showarticletitle{Extensions of recurrent neural network language
  model}. In \bibinfo{booktitle}{\emph{Acoustics, Speech and Signal Processing
  (ICASSP), 2011 IEEE International Conference on}}. IEEE,
  \bibinfo{pages}{5528--5531}.
\newblock


\bibitem[\protect\citeauthoryear{Neil, Pfeiffer, and Liu}{Neil
  et~al\mbox{.}}{2016}]%
        {neil-etal2016}
\bibfield{author}{\bibinfo{person}{Daniel Neil}, \bibinfo{person}{Michael
  Pfeiffer}, {and} \bibinfo{person}{Shih-Chii Liu}.}
  \bibinfo{year}{2016}\natexlab{}.
\newblock \showarticletitle{Phased lstm: Accelerating recurrent network
  training for long or event-based sequences}. In
  \bibinfo{booktitle}{\emph{Advances in Neural Information Processing
  Systems}}. \bibinfo{pages}{3882--3890}.
\newblock


\bibitem[\protect\citeauthoryear{Palangi, Deng, Shen, Gao, He, Chen, Song, and
  Ward}{Palangi et~al\mbox{.}}{2016}]%
        {palangi-etal2016}
\bibfield{author}{\bibinfo{person}{Hamid Palangi}, \bibinfo{person}{Li Deng},
  \bibinfo{person}{Yelong Shen}, \bibinfo{person}{Jianfeng Gao},
  \bibinfo{person}{Xiaodong He}, \bibinfo{person}{Jianshu Chen},
  \bibinfo{person}{Xinying Song}, {and} \bibinfo{person}{Rabab Ward}.}
  \bibinfo{year}{2016}\natexlab{}.
\newblock \showarticletitle{Deep sentence embedding using long short-term
  memory networks: Analysis and application to information retrieval}.
\newblock \bibinfo{journal}{\emph{IEEE/ACM Transactions on Audio, Speech and
  Language Processing (TASLP)}} \bibinfo{volume}{24}, \bibinfo{number}{4}
  (\bibinfo{year}{2016}), \bibinfo{pages}{694--707}.
\newblock


\bibitem[\protect\citeauthoryear{Rumelhart, Hinton, and Williams}{Rumelhart
  et~al\mbox{.}}{1986a}]%
        {rumelhart1986learning}
\bibfield{author}{\bibinfo{person}{David~E Rumelhart},
  \bibinfo{person}{Geoffrey~E Hinton}, {and} \bibinfo{person}{Ronald~J
  Williams}.} \bibinfo{year}{1986}\natexlab{a}.
\newblock \showarticletitle{Learning representations by back-propagating
  errors}.
\newblock \bibinfo{journal}{\emph{nature}} \bibinfo{volume}{323},
  \bibinfo{number}{6088} (\bibinfo{year}{1986}), \bibinfo{pages}{533}.
\newblock


\bibitem[\protect\citeauthoryear{Rumelhart, Hinton, and Williams}{Rumelhart
  et~al\mbox{.}}{1986b}]%
        {rumelhart-etal1986}
\bibfield{author}{\bibinfo{person}{David~E Rumelhart},
  \bibinfo{person}{Geoffrey~E Hinton}, {and} \bibinfo{person}{Ronald~J
  Williams}.} \bibinfo{year}{1986}\natexlab{b}.
\newblock \showarticletitle{Learning representations by back-propagating
  errors}.
\newblock \bibinfo{journal}{\emph{nature}} \bibinfo{volume}{323},
  \bibinfo{number}{6088} (\bibinfo{year}{1986}), \bibinfo{pages}{533}.
\newblock


\bibitem[\protect\citeauthoryear{Schuster and Paliwal}{Schuster and
  Paliwal}{1997}]%
        {schuster-etal1997}
\bibfield{author}{\bibinfo{person}{Mike Schuster} {and}
  \bibinfo{person}{Kuldip~K Paliwal}.} \bibinfo{year}{1997}\natexlab{}.
\newblock \showarticletitle{Bidirectional recurrent neural networks}.
\newblock \bibinfo{journal}{\emph{IEEE Transactions on Signal Processing}}
  \bibinfo{volume}{45}, \bibinfo{number}{11} (\bibinfo{year}{1997}),
  \bibinfo{pages}{2673--2681}.
\newblock


\bibitem[\protect\citeauthoryear{Soltau, Liao, and Sak}{Soltau
  et~al\mbox{.}}{2016}]%
        {soltau-etal2016}
\bibfield{author}{\bibinfo{person}{Hagen Soltau}, \bibinfo{person}{Hank Liao},
  {and} \bibinfo{person}{Hasim Sak}.} \bibinfo{year}{2016}\natexlab{}.
\newblock \showarticletitle{Neural speech recognizer: Acoustic-to-word LSTM
  model for large vocabulary speech recognition}.
\newblock \bibinfo{journal}{\emph{arXiv preprint arXiv:1610.09975}}
  (\bibinfo{year}{2016}).
\newblock


\bibitem[\protect\citeauthoryear{Sutskever, Martens, and Hinton}{Sutskever
  et~al\mbox{.}}{2011}]%
        {sutskever2011generating}
\bibfield{author}{\bibinfo{person}{Ilya Sutskever}, \bibinfo{person}{James
  Martens}, {and} \bibinfo{person}{Geoffrey~E Hinton}.}
  \bibinfo{year}{2011}\natexlab{}.
\newblock \showarticletitle{Generating text with recurrent neural networks}. In
  \bibinfo{booktitle}{\emph{Proceedings of the 28th International Conference on
  Machine Learning (ICML-11)}}. \bibinfo{pages}{1017--1024}.
\newblock


\bibitem[\protect\citeauthoryear{Sutskever, Vinyals, and Le}{Sutskever
  et~al\mbox{.}}{2014}]%
        {sutskever-etal2014}
\bibfield{author}{\bibinfo{person}{Ilya Sutskever}, \bibinfo{person}{Oriol
  Vinyals}, {and} \bibinfo{person}{Quoc~V Le}.}
  \bibinfo{year}{2014}\natexlab{}.
\newblock \showarticletitle{Sequence to sequence learning with neural
  networks}. In \bibinfo{booktitle}{\emph{Advances in neural information
  processing systems}}. \bibinfo{pages}{3104--3112}.
\newblock


\bibitem[\protect\citeauthoryear{Tang, Hu, and Liu}{Tang et~al\mbox{.}}{2013}]%
        {tang-etal2013}
\bibfield{author}{\bibinfo{person}{Jiliang Tang}, \bibinfo{person}{Xia Hu},
  {and} \bibinfo{person}{Huan Liu}.} \bibinfo{year}{2013}\natexlab{}.
\newblock \showarticletitle{Social recommendation: a review}.
\newblock \bibinfo{journal}{\emph{Social Network Analysis and Mining}}
  \bibinfo{volume}{3}, \bibinfo{number}{4} (\bibinfo{year}{2013}),
  \bibinfo{pages}{1113--1133}.
\newblock


\bibitem[\protect\citeauthoryear{Tang and Liu}{Tang and Liu}{2012}]%
        {tang-etal2012}
\bibfield{author}{\bibinfo{person}{Jiliang Tang} {and} \bibinfo{person}{Huan
  Liu}.} \bibinfo{year}{2012}\natexlab{}.
\newblock \showarticletitle{Unsupervised feature selection for linked social
  media data}. In \bibinfo{booktitle}{\emph{Proceedings of the 18th ACM SIGKDD
  international conference on Knowledge discovery and data mining}}. ACM,
  \bibinfo{pages}{904--912}.
\newblock


\bibitem[\protect\citeauthoryear{Tang, Qu, Wang, Zhang, Yan, and Mei}{Tang
  et~al\mbox{.}}{2015}]%
        {tang-etal2015}
\bibfield{author}{\bibinfo{person}{Jian Tang}, \bibinfo{person}{Meng Qu},
  \bibinfo{person}{Mingzhe Wang}, \bibinfo{person}{Ming Zhang},
  \bibinfo{person}{Jun Yan}, {and} \bibinfo{person}{Qiaozhu Mei}.}
  \bibinfo{year}{2015}\natexlab{}.
\newblock \showarticletitle{Line: Large-scale information network embedding}.
  In \bibinfo{booktitle}{\emph{Proceedings of the 24th International Conference
  on World Wide Web}}. International World Wide Web Conferences Steering
  Committee, \bibinfo{pages}{1067--1077}.
\newblock


\bibitem[\protect\citeauthoryear{Tang, Zhang, Yao, Li, Zhang, and Su}{Tang
  et~al\mbox{.}}{2008}]%
        {Tang-etal2008}
\bibfield{author}{\bibinfo{person}{Jie Tang}, \bibinfo{person}{Jing Zhang},
  \bibinfo{person}{Limin Yao}, \bibinfo{person}{Juanzi Li}, \bibinfo{person}{Li
  Zhang}, {and} \bibinfo{person}{Zhong Su}.} \bibinfo{year}{2008}\natexlab{}.
\newblock \showarticletitle{ArnetMiner: Extraction and Mining of Academic
  Social Networks}. In \bibinfo{booktitle}{\emph{KDD'08}}.
  \bibinfo{pages}{990--998}.
\newblock


\bibitem[\protect\citeauthoryear{Wang, Wei, Liu, Zhou, and Zhang}{Wang
  et~al\mbox{.}}{2011}]%
        {wang-etal2011}
\bibfield{author}{\bibinfo{person}{Xiaolong Wang}, \bibinfo{person}{Furu Wei},
  \bibinfo{person}{Xiaohua Liu}, \bibinfo{person}{Ming Zhou}, {and}
  \bibinfo{person}{Ming Zhang}.} \bibinfo{year}{2011}\natexlab{}.
\newblock \showarticletitle{Topic sentiment analysis in twitter: a graph-based
  hashtag sentiment classification approach}. In
  \bibinfo{booktitle}{\emph{Proceedings of the 20th ACM international
  conference on Information and knowledge management}}. ACM,
  \bibinfo{pages}{1031--1040}.
\newblock


\bibitem[\protect\citeauthoryear{Wang, Derr, Yin, and Tang}{Wang
  et~al\mbox{.}}{2017}]%
        {wang-etal2017}
\bibfield{author}{\bibinfo{person}{Zhiwei Wang}, \bibinfo{person}{Tyler Derr},
  \bibinfo{person}{Dawei Yin}, {and} \bibinfo{person}{Jiliang Tang}.}
  \bibinfo{year}{2017}\natexlab{}.
\newblock \showarticletitle{Understanding and Predicting Weight Loss with
  Mobile Social Networking Data}. In \bibinfo{booktitle}{\emph{Proceedings of
  the 2017 ACM on Conference on Information and Knowledge Management}}. ACM,
  \bibinfo{pages}{1269--1278}.
\newblock


\bibitem[\protect\citeauthoryear{Wu, Wang, Liu, and Liu}{Wu
  et~al\mbox{.}}{2016}]%
        {wu-etal2016}
\bibfield{author}{\bibinfo{person}{Caihua Wu}, \bibinfo{person}{Junwei Wang},
  \bibinfo{person}{Juntao Liu}, {and} \bibinfo{person}{Wenyu Liu}.}
  \bibinfo{year}{2016}\natexlab{}.
\newblock \showarticletitle{Recurrent neural network based recommendation for
  time heterogeneous feedback}.
\newblock \bibinfo{journal}{\emph{Knowledge-Based Systems}}
  \bibinfo{volume}{109} (\bibinfo{year}{2016}), \bibinfo{pages}{90--103}.
\newblock


\bibitem[\protect\citeauthoryear{Yang, Yang, Dyer, He, Smola, and Hovy}{Yang
  et~al\mbox{.}}{2016}]%
        {yang-etal2016}
\bibfield{author}{\bibinfo{person}{Zichao Yang}, \bibinfo{person}{Diyi Yang},
  \bibinfo{person}{Chris Dyer}, \bibinfo{person}{Xiaodong He},
  \bibinfo{person}{Alex Smola}, {and} \bibinfo{person}{Eduard Hovy}.}
  \bibinfo{year}{2016}\natexlab{}.
\newblock \showarticletitle{Hierarchical attention networks for document
  classification}. In \bibinfo{booktitle}{\emph{Proceedings of the 2016
  Conference of the North American Chapter of the Association for Computational
  Linguistics: Human Language Technologies}}. \bibinfo{pages}{1480--1489}.
\newblock


\bibitem[\protect\citeauthoryear{Zhou, Ding, Tang, and Yin}{Zhou
  et~al\mbox{.}}{2018}]%
        {zhou-etal2018}
\bibfield{author}{\bibinfo{person}{Meizi Zhou}, \bibinfo{person}{Zhuoye Ding},
  \bibinfo{person}{Jiliang Tang}, {and} \bibinfo{person}{Dawei Yin}.}
  \bibinfo{year}{2018}\natexlab{}.
\newblock \showarticletitle{Micro behaviors: A new perspective in e-commerce
  recommender systems}. In \bibinfo{booktitle}{\emph{Proceedings of the
  Eleventh ACM International Conference on Web Search and Data Mining}}. ACM,
  \bibinfo{pages}{727--735}.
\newblock


\end{thebibliography}

\end{document}